\documentclass[letterpaper, 10 pt, conference]{ieeeconf}  

\usepackage{graphicx}
\usepackage{listings}
\IEEEoverridecommandlockouts                              

\usepackage{hyperref}
\usepackage{xcolor}
\usepackage{mars}

\title{\LARGE \bf
High-Quality, ROS Compatible Video Encoding and\\ Decoding for High-Definition Datasets
}

\author{Jian Li, Bowen Xu, S\"oren Schwertfeger
\thanks{This work was supported by the Science and Technology Commission of Shanghai Municipality (STCSM), project 22JC1410700 "Evaluation of real-time localization and mapping algorithms for intelligent robots". This work has also been partially funded by the Shanghai Frontiers Science Center of Human-centered Artificial Intelligence. The experiments of this work were supported by the core facility Platform of Computer Science and Communication, SIST, ShanghaiTech University.
}
\thanks{$^{*}$All authors are with the Key Laboratory of Intelligent Perception and Human-Machine Collaboration -- ShanghaiTech University, Ministry of Education, China;
{\tt\small soerensch@shanghaitech.edu.cn}}%
}

\begin{document}

%
%


 \marsPublishedIn{Accepted for:} 		

\marsVenue{IEEE International Conference on Robotics and Biomimetics(ROBIO) 2024}

\marsYear{2024}

\marsPlainAutors{Jian Li, Bowen Xu, S\"oren Schwertfeger}


\marsMakeCitation{High-Quality, ROS Compatible Video Encoding and Decoding for High-Definition Datasets}{IEEE Press}


\marsIEEE{}


\makeMARStitle

\title{\LARGE \bf
	 High-Quality, ROS Compatible Video Encoding and\\ Decoding for High-Definition Datasets
}

\author{Jian Li$^{*}$, Bowen Xu, S\"oren Schwertfeger
	\thanks{This work was supported by the Science and Technology Commission of Shanghai Municipality (STCSM), project 22JC1410700 "Evaluation of real-time localization and mapping algorithms for intelligent robots". This work has also been partially funded by the Shanghai Frontiers Science Center of Human-centered Artificial Intelligence. The experiments of this work were supported by the core facility Platform of Computer Science and Communication, SIST, ShanghaiTech University.
	}
	\thanks{$^{*}$All authors are with the Key Laboratory of Intelligent Perception and Human-Machine Collaboration -- ShanghaiTech University, Ministry of Education, China;
		{\tt\small soerensch@shanghaitech.edu.cn}}%
}

\maketitle
\thispagestyle{empty}
\pagestyle{empty}

\begin{abstract}
Robotic datasets are important for scientific benchmarking and developing algorithms, for example for Simultaneous Localization and Mapping (SLAM). Modern robotic datasets feature video data of high resolution and high framerates. Storing and sharing those datasets becomes thus very costly, especially if more than one camera is used for the datasets. It is thus essential to store this video data in a compressed format. This paper investigates the use of modern video encoders for robotic datasets. We provide a software that can replay mp4 videos within ROS 1 and ROS 2 frameworks, supporting the synchronized playback in simulated time. Furthermore, the paper evaluates different encoders and their settings to find optimal configurations in terms of resulting size, quality and encoding time. Through this work we show that it is possible to store and share even highest quality video datasets within reasonable storage constraints. 
\end{abstract}

\section{INTRODUCTION}

Datasets are an important tool for research and development in mobile robotics. For example, for evaluating Simultaneous Localization and Mapping (SLAM) algorithms \cite{cadena2016past}, datasets play an important role \cite{liu2024benchmarking, yang2023slam}. Sensor technology is advancing rapidly, such that better camera sensors with higher resolutions and higher frame rates are available for use in robotic systems. While older datasets like KITTI from 2013 \cite{Geiger2012CVPR} use 1.4MP cameras with a frame rate of 10Hz, later datasets offer for example 5MP images with 10 Hz \cite{chen2020advanced}. 

Our most recent mapping robot for collecting datasets features five pairs of stereo cameras, facing front, left, right, back and up, each with 5MP \cite{xu2024shanghaitech}, and a 16 node "Cluster on Wheels" \cite{yang2022cluster} to collect this data with 60Hz. Additionally, that robot features other vision sensors (e.g. a 6-lens Ladybug 5+, stereo infrared, RGB-D, event camers) as well as 7 LiDARs. The datasets collected with this robot are valuable tools to evaluate the performance of SLAM algorithms. 

It is especially interesting to use this high-end data to then simulate lower quality data by resizing the images to get lower resolutions or dropping frames to simulate lower frame rates. This way, one can investigate in detail the effects of resolution and frame rate vs. accuracy, CPU and memory consumption. We have developed the SLAM Hive system \cite{liu2024benchmarking, yang2023slam} to do such benchmarking automatically in the cloud and provide an easy to use web interface for analysis of the data\footnote{SLAM Hive Demo: \url{https://slam-hive.net/}}. 

As discussed, high resolution and high-quality datasets are valuable for evaluating SLAM. The problem is with this data is, that it is really big. As can be seen in Table \ref{tab:compression result}, the 11 5MP cameras with just 30Hz frame rate of the initial ShanghaiTech Mapping Robot Outdoor Dataset \cite{xu2024shanghaitech} has a raw data size of more than 2TB, the RGB size would be more than 7TB and even not-so-great quality 90 JPG would still be 446GB that would needed to be shared with the community. This is intractable in terms of transmission time over the internet, but also in terms of storage space for most computers. Therefore, the goal of this paper is to investigate how better compression rates can be achieved, while still maintaining excellent image quality, reasonable encoding times and small size. In order for the compressed dataset to be useful as a reference dataset for SLAM research and benchmarking, a truly excellent video quality has to be maintained. As reported in \cite{chan2023influence, gummadi2023correlating}, compression artefacts can in fact reduce the applicability of video streams to various applications such as object detection. Specifically, we do not need to encode the videos in real-time, but will do the compression in post-processing, because there are not enough computation resources even on a 16-node cluster to do the de-bayering and encoding in high-quality in real time. In the end our experiments will justify our choice of encoder and settings, which enables us to share the above mentioned video dataset with a size of just 43GB.

The contributions of this paper are:
\begin{itemize}
	\item Investigation of different high-quality compression/ encoding options for video data;
	\item Experimental evaluations of the options based on a high-resolution, high-frame rate test video;
	\item Development and open-sourcing \footnote{\url{https://github.com/STAR-Center/ros_mp4}} of a python tool to replay MP4 videos with yaml metadata, including support for synchronized replay with simulated time;
	\item Evaluation of source frame rate effects on encoding.
\end{itemize}

The rest of the paper is organized as follows: Section \ref{sec:background} introduces the background and related work. Section \ref{sec:playback} introduces our software framework for encoding and decoding the videos within the ROS 1 and ROS 2 middle-ware. Extensive evaluations regarding encoding algorithms and their settings vs. quality, encoding time and video size and further experiments are performed in Section \ref{sec:experiments}  and conclusions are drawn in Section \ref{sec:conclusions}.

\section{BACKGROUND}
\label{sec:background}

\begin{figure}[t]
	\centering
	\includegraphics[width=\columnwidth]{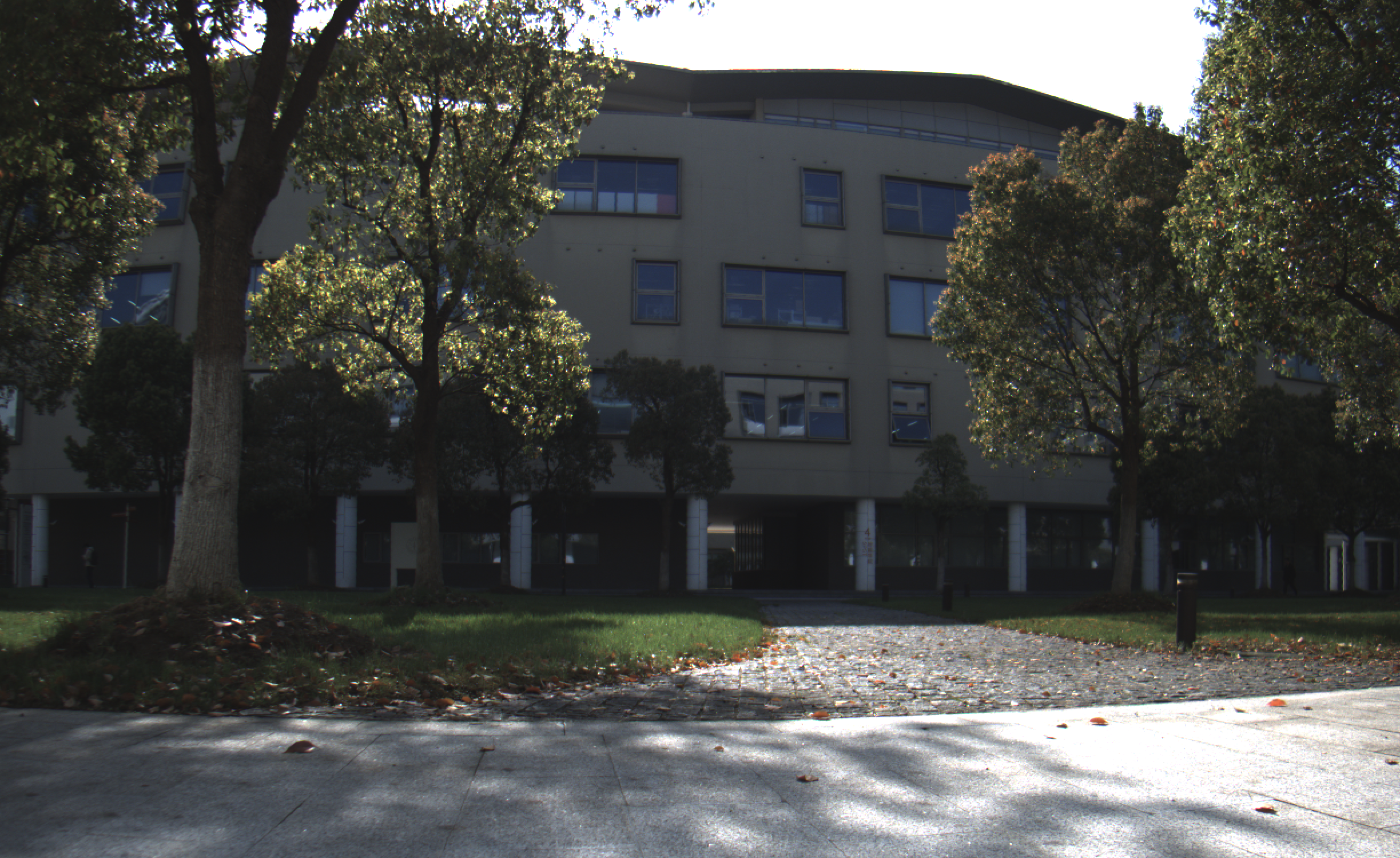}
	\caption{One frame from our test clip}
	\label{fig:dataset-clip}
	
\end{figure}

When collecting video datasets, several options for storage come to mind. In this paper we only consider color video data, as this is the most prevalent and information rich video format. High-quality computer vision cameras, such as the Grasshopper3 (GS3-U3-51S5C-C), provide raw data, that is bayered \cite{bayer1976color}. This means, that there are four channels of red, green (2x) and blue gray-scale values provided, typically each using one byte. After de-bayering, the image size is tripled, since every pixel now gets an RGB or YUV444 color value, interpolated over the surrounding, colored pixels, each using 3 bytes. Given the 5,013,504 pixels (2048x2448) of the 5MP camera and a 60Hz update rate, this results in 902MB per second or 3.3 TB per hour - for one camera stream. Thus compression is needed. One example frame is shown in Fig. \ref{fig:dataset-clip}.

Traditionally, robotics is concerned with real-time video compression for teleoperation \cite{ramil2018real}, robot-to-robot communication \cite{taguchi2008video} or transmission of range data \cite{nenci2014effective}. But the aim for this paper are datasets with no visible degradation in image quality. Lossless image compression approaches like PNG do not help much with real camera data, since the camera noise results in most pixels having different values even if they look similar. The next best option is to use JPEG. In Table \ref{tab:compression result} we show that even very low quality jpeg values such as 75 result big file sizes. Thus real video encoders promise better video quality and smaller size.

\subsection{ROS and Video}

Robotic datasets often use ROS, specifically ROS bags to store the data. Video data in ROS 1 is stored in image messages, one for each frame. The sensor\_msgs/Image message type provides different encoding options, such as bayer\_GBRG8, rgba8, yuv422 and OpenCV types  \footnote{\url{http://docs.ros.org/en/indigo/api/sensor_msgs/html/image__encodings_8h_source.html}}. The type sensor\_msgs/CompressedImage, on the other hand, supports PNG, JPG and h264. ROS 2 also supports newer codecs, such as h265 \footnote{\url{https://github.com/ros-misc-utilities/ffmpeg_image_transport}}. 

An important feature of ROS is the time-synchronized playback with simulated time. For that ROS will take the time stamps of one bagfile and publish that time on the topic /clock. Other rosbag play instances will listen to that simulated time, if the parameter use\_sim\_time is set to true beforehand, and play their messages at the according, simulated times, thus achiving a synchronized playback of multiple bagfiles. 

In this project we decided to encode the video data of the dataset as mp4 files. This has several advantages: 1) Users can inspect the data very easily, by just playing them in their favorite video player (e.g. VLC). 2) We can use the newest encoders and are not held back by compatibility issues regarding provided encoders of the compressed image format. 3) Our playback tools work with ROS 1 and with ROS 2, so it is very easy to use the video dataset with either software.

To support the functionality similar to rosbag play, our software uses metadata, specifially the time stamps of the image frames, stored in an external yaml file. Using this we can replay the videos in ROS 1 and ROS 2 just as if they came from a rosbag, including support for time-synchronize playback via simulated time.

Next the different options for video codecs are presented.
 
\subsubsection{H.264/AVC}

In 2004 the codec H.264/AVC was introduced \cite{tamhankar2003overview}. It has numerous innovative features such as multi-reference frame motion compensation, variable block size motion compensation, and intra-frame prediction encoding.


\subsubsection{H.265/HEVC}
The H.265/HEVC  codec was introduced in 2013 \cite{sullivan2012overview,pourazad2012hevc}. It further optimizes predictive coding techniques based on the inheritance of H.264/AVC, including more refined intra-frame and inter-frame prediction\cite{patel2015review}, larger coding units\cite{kim2012block}, and higher precision motion vectors\cite{lin2013motion}, significantly improving compression efficiency. 

\newcommand{\jlwdithA}{0.15\linewidth}

\newcommand{\jlwidthB}{1\linewidth}

\begin{table*}[ht]
	\centering
	\caption{Visual comparison of different VMAF scored of detail of I-frame 1284 of the test video.} 
	\label{tab:av1-vision}
	\begin{tabular}{c|c|c|c|c|c}
		
		\parbox[t]{\jlwdithA}{\centering \includegraphics[width=\jlwidthB]{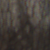} \\ Ground Truth} 
		& \parbox[t]{\jlwdithA}{\centering \includegraphics[width=\jlwidthB]{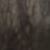} \\ VMAF:  98.68}
		& \parbox[t]{\jlwdithA}{\centering \includegraphics[width=\jlwidthB]{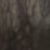} \\  VMAF:  98.82}
		& \parbox[t]{\jlwdithA}{\centering \includegraphics[width=\jlwidthB]{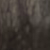} \\  VMAF:  98.95}
		& \parbox[t]{\jlwdithA}{\centering \includegraphics[width=\jlwidthB]{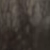} \\  VMAF:  99.16}
		& \parbox[t]{\jlwdithA}{\centering \includegraphics[width=\jlwidthB]{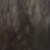} \\  VMAF:  99.29\vspace{2mm}}
		\\
		\parbox[t]{\jlwdithA}{\centering \includegraphics[width=\jlwidthB]{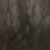} \\  VMAF:  99.33}
		& \parbox[t]{\jlwdithA}{\centering \includegraphics[width=\jlwidthB]{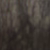} \\   VMAF:  99.40} 
		& \parbox[t]{\jlwdithA}{\centering \includegraphics[width=\jlwidthB]{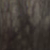} \\  VMAF:  99.44}
		& \parbox[t]{\jlwdithA}{\centering \includegraphics[width=\jlwidthB]{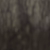} \\  VMAF:  99.50} 		 
		& \parbox[t]{\jlwdithA}{\centering \includegraphics[width=\jlwidthB]{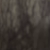} \\   VMAF:  99.52} 
		& \parbox[t]{\jlwdithA}{\centering \includegraphics[width=\jlwidthB]{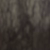} \\   VMAF:  99.55} 
		\\ 
\end{tabular} \end{table*}

\subsubsection{AV1}
Following H.264/AVC and H.265/HEVC, the AV1\cite{han2021technical} encoding standard was introduced in 2018. It features high compression efficiency and is open-source. AV1 upgrades the granularity of intra-frame directional prediction and boasts a more powerful inter-frame encoder\cite{joshi2019loop}, significantly enhancing the accuracy of compound prediction.


\subsection{FFmpeg}
FFmpeg\cite{tomar2006converting} is an open-source multimedia framework supporting  numerous codecs, including H.264, H.265, and AV1.
%
%
Among the key parameters in FFmpeg's command line, -CRF (Constant Rate Factor) and -PRESET are crucial for controlling compression quality. CRF is a "constant quality" encoding mode that dynamically adjusts quantization parameters (QP) across frames to maintain a consistent visual quality. Users can adjust the quality of the output video by specifying a CRF value between 0 and 51, where a lower value corresponds to higher video quality and a larger file size.The -PRESET parameter balances encoding speed and compression efficiency, offering options ranging from "ultrafast" to "placebo", allowing users to choose the appropriate encoding speed based on specific application scenarios and hardware performance.

\subsection{Image Quality Measurement}

To ensure the visual quality of videos, conducting detailed subjective and objective evaluations of video quality is crucial. VMAF (Video Multi-Method Assessment Fusion) is a  video quality assessment tool jointly developed by Netflix and the academic community \cite{rassool2017vmaf}. 

VMAF measures video quality through integrating multiple visual quality indicators, generating objective scores that highly align with human visual perception. Based on the nuSvr algorithm of SVM, VMAF first extracts key visual features in the evaluation process, including but not limited to Visual Information Fidelity (VIF), Detail Loss Metric (DLM), and Temporal Information (TI), comprehensively considering the spatial and temporal dimensions of the video. Then, VMAF utilizes a pre-trained SVM model to combine these features with human subjective scores, assigning different weights to each video feature. A score is generated for each frame, and the final score of the video is calculated using the mean of those values. 

Table~\ref{tab:av1-vision} shows details of an I-frame of our test video encoded with different parameters, thus resulting in different VMAF scores. There are sill some differences in the frames visible for VMAF scores below 99.50. But for scores higher than that no difference can be made out with the naked eye. Therefore we chose 99.50 as our target VMAF score for highest-quality video encoding. 

\section{ROS MP4 Encoding and Playback Framework}
\label{sec:playback}

In this section, we will delve into the detailed steps of compressing rosbag and playback. The camera image messages recorded in rosbag are essentially a series of consecutive image frames with timestamps and other metadata, which is quite similar to the characteristics of video streams. 
The first step is to separate the image frames from their associated timestamps and other metadata. The camera images recorded in rosbag may not be directly stored in common BGR formats, but may use formats like "bayer\_gbrg8" or may have already undergone some form of compression during recording. In such cases, necessary format conversion and preprocessing may be required. Once the image frames are extracted, we can utilize  FFmpeg to compress these image frames into the desired video format.
\subsection{Yaml File}
Apart from image frames, other metadata (such as timestamps, frame IDs, intrinsic parameters, etc.) also need to be properly preserved to ensure that we can fully restore all the original information from the rosbag messages. To achieve this, we use the YAML file format to store these metadata.  For typical visual rosbag datasets, there are two key topics: "image\_raw" (or a similar name) for storing raw image data, and "camera\_info" containing camera calibration and distortion information related to each frame. After analyzing the information structure of these two topics, we find that the information in the "camera\_info" topic actually contains some key information corresponding to the image frames in the "image\_raw" topic, such as frame\_id and timestamp. Therefore, we define the structure of the YAML file as follows:


\begin{figure}[h]
    \centering
    \includegraphics[width=\columnwidth]{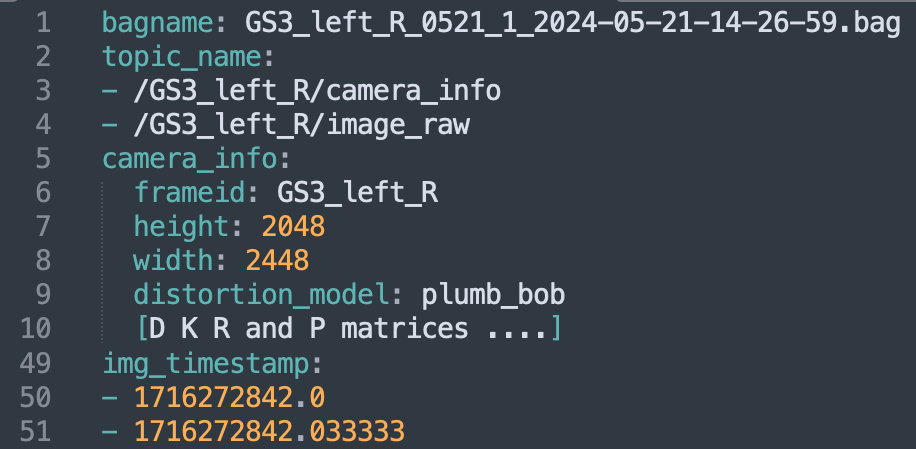}
    \label{fig:yaml}
\end{figure}

%


\begin{table}[t]
	\centering
	\caption{x264 test data}
	\label{tab:libx264-table}
	\begin{tabular}{rccccr} 
		\hline
		Time (s) & CRF & Preset & Bitrate (Kbps) & Size (MB) & VMAF \\
		\hline
		93.06 & 19 & medium & 46,561 & 333 & 98.78 \\
		271.86 & 19 & slower & 45,341 & 324 & 98.93 \\
		170.82 & 19 & slow & 46,851 & 335 & 98.94 \\
		502.79 & 19 & veryslow & 44,245 & 316 & 98.97 \\
		1954.65 & 19 & placebo & 46,401 & 332 & 99.10 \\
		96.74 & 18 & medium & 55,101 & 394 & 99.14 \\
		288.28 & 18 & slower & 53,413 & 382 & 99.22 \\
		181.70 & 18 & slow & 5,5354 & 396 & 99.22 \\
		528.76 & 18 & veryslow & 52,149 & 373 & 99.24 \\
		1994.10 & 18 & placebo & 54,711 & 391 & 99.32 \\
		100.81 & 17 & medium & 65,651 & 470 & 99.33 \\
		192.92 & 17 & slow & 65,823 & 471 & 99.37 \\
		308.21 & 17 & slower & 63,285 & 453 & 99.38 \\
		560.15 & 17 & veryslow & 61,941 & 443 & 99.38 \\
		105.06 & 16 & medium & 78,577 & 562 & 99.43 \\
		2041.67 & 17 & placebo & 64,999 & 465 & 99.44 \\
		206.21 & 16 & slow & 78,642 & 562 & 99.46 \\
		592.37 & 16 & veryslow & 73,964 & 529 & 99.47 \\
		328.06 & 16 & slower & 75,361 & 539 & 99.47 \\
		2087.55 & 16 & placebo & 77,654 & 555 & 99.49 \\
		\hline
	\end{tabular}
	\vspace{-5mm}
\end{table}

\begin{table}[b]
	{
	\centering
	\caption{60 second test video experiment: svtav1 with VMAF scores greater  than 99.45 and compression times less than 1000 seconds. Ultimately we chose the settings in bold.}
	\label{tab:Full size}
   }
	\begin{tabular}{ccccccc} 
		\hline
		CRF & Pre.$^1$ & Bitrate & Size& VMAF & Full Size& Full Enc.$^2$\\
		& & (Kbps) & (MB) & &(GB) & Time (days)\\
		\hline
		24 & 5 & 44,298 & 317 & 99.46 & 557 & 7.99\\
		21 & 7 & 52,904 & 378 & 99.47 & 665 & 2.65\\
		23 & 6 & 47,116 & 337 & 99.47 & 592 & 4.94\\
		17 & 8 & 72,933 & 522 & 99.49 & 917& 1.61\\
		\textbf{22} & \textbf{6} & \textbf{49,812} & \textbf{356} & \textbf{99.50} & \textbf{626} & \textbf{4.87}\\
		22 & 5 & 49,608 & 355& 99.51 & 624 & 8.12 \\
		16 & 8 & 78,263 & 560 & 99.51 & 984 & 1.62 \\
		19 & 7 & 62,127 & 444 & 99.52 & 781 & 2.72 \\
		21 & 6 & 52,941 & 379 & 99.52 & 666 & 4.95 \\
		21 & 5 & 52,477 & 375 & 99.53 & 660 & 8.20 \\
		20 & 6 & 57,046 & 408 & 99.54 & 717 & 5.00 \\
		20 & 5 & 55,718 & 399 & 99.54 & 701 & 8.28 \\
		17 & 7 & 71,748 & 513 & 99.55 & 902 & 2.81 \\
		19 & 6 & 61,599 & 441 & 99.55 & 774 & 5.07 \\
		16 & 7 & 77,582 & 555 & 99.56 & 975 & 2.88 \\
		19 & 5 & 60,055 & 430 & 99.56 & 755 & 8.41 \\
		18 & 5 & 64,473 & 461 & 99.57 & 811 & 8.51 \\
		17 & 6 & 71,216 & 509 & 99.57 & 895& 5.20 \\
		16 & 5 & 73,853 & 528 & 99.58 & 929 & 8.73 \\
		\hline
	\end{tabular}
    $^1$ Preset \hspace{5mm} $^2$ Full Encoding Time (days)
\end{table}

\subsection{Play Back}
\begin{figure}[b]
	\centering
	\includegraphics[width=\columnwidth]{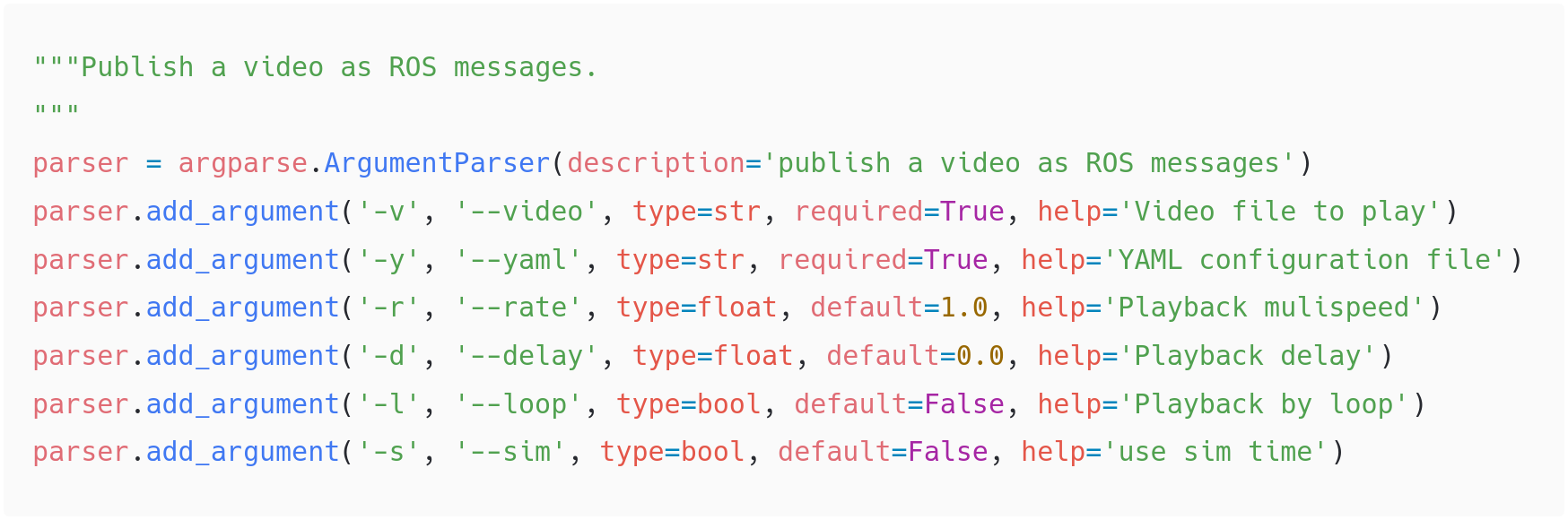}
	\caption{Playback tool options.}
	\label{fig:tool}
	\vspace{-9mm}
\end{figure}

To play back this compressed dataset composed of YAML file and video, we have developed a powerful script. This script not only integrates and publishes topic information corresponding to the original rosbag dataset, but also retains many of the same features as playing a rosbag. By parsing the timestamp information in the YAML file and utilizing ROS's clock mechanism, our script can accurately simulate the playback time of the dataset in a real-world environment, ensuring that the content of the published topics remains consistent with the original rosbag playback in terms of both timing and content. This high level of accuracy provides users with an experience almost identical to directly playing a rosbag. Fig. \ref{fig:tool} shows the different playback options available. 


\begin{table}[t]
	\centering
	\caption{x265 test data}
	\label{tab:libx265-table}
	\begin{tabular}{rccccr} 
		\hline
		Time (s) & CRF & Preset & Bitrate (Kbps) & Size (MB) & VMAF \\
		\hline
		5808.90 & 20 & veryslow & 43,602 & 312 & 99.23 \\
		295.81 & 17 & medium & 64,506 & 461 & 99.24 \\
		724.26 & 19 & slow & 49,860 & 357 & 99.30 \\
		10772.59 & 20 & placebo & 46,148 & 330 & 99.32 \\
		3487.48 & 19 & slower & 51,473 & 368 & 99.37 \\
		6205.14 & 19 & veryslow & 51,479 & 368 & 99.38 \\
		313.50 & 16 & medium & 76,809 & 549 & 99.40 \\
		749.04 & 18 & slow & 59,219 & 424 & 99.43 \\
		11293.54 & 19 & placebo & 54,470 & 390 & 99.44 \\
		3722.43 & 18 & slower & 60,926 & 436 & 99.47 \\
		6608.98 & 18 & veryslow & 60,942 & 435 & 99.48 \\
		785.58 & 17 & slow & 70,482 & 504 & 99.50 \\
		11774.43 & 18 & placebo & 64,409 & 461 & 99.51 \\
		3977.30 & 17 & slower & 72,232 & 517 & 99.52 \\
		7041.76 & 17 & veryslow & 72,240 & 517 & 99.53 \\
		832.83 & 16 & slow & 83,992 & 600 & 99.55 \\
		12337.12 & 17 & placebo & 76,237 & 545 & 99.55 \\
		4260.37 & 16 & slower & 85,720 & 613 & 99.55 \\
		7571.67 & 16 & veryslow & 85,716 & 613 & 99.56 \\
		13015.65 & 16 & placebo & 90,273 & 646 & 99.57 \\
		\hline
	\end{tabular}
		\vspace{-5mm}
\end{table}

\section{Encoding Experiments}
\label{sec:experiments}

To determine the optimal compression combination for video quality and compression efficiency for the new dataset structure, we carefully selected a 1-minute clip from the recorded dataset that contains both static and dynamic scenes as a test sample. This clip combines a 5-second static outdoor scene and a 55-second dynamic outdoor scene, which can represent the typical characteristics of the dataset. The video was recorded on the ShaghaiTech Mapping Robot with a Grasshopper3 (GS3-U3-51S5C-C) camera with 2048x2448 pixel at 60Hz frame rate.

For evaluation we utilize the 60 second test video, but then extrapolate the results to a dataset of 14 such video streams each 2 hours long, as this is the expected dimension of our future ShanghaiTech Mapping Robot Datasets. The robot features 11 such datastreams at full frame rate plus the LadyBug5+ camera with 6 streams at half the frame rate, so in total around $11 + 6/2 = 14$. The results help us to understand the computation and storage demands of a dataset compressed with the according settings.

\begin{figure}[t]
	\centering
	\includegraphics[width=\columnwidth]{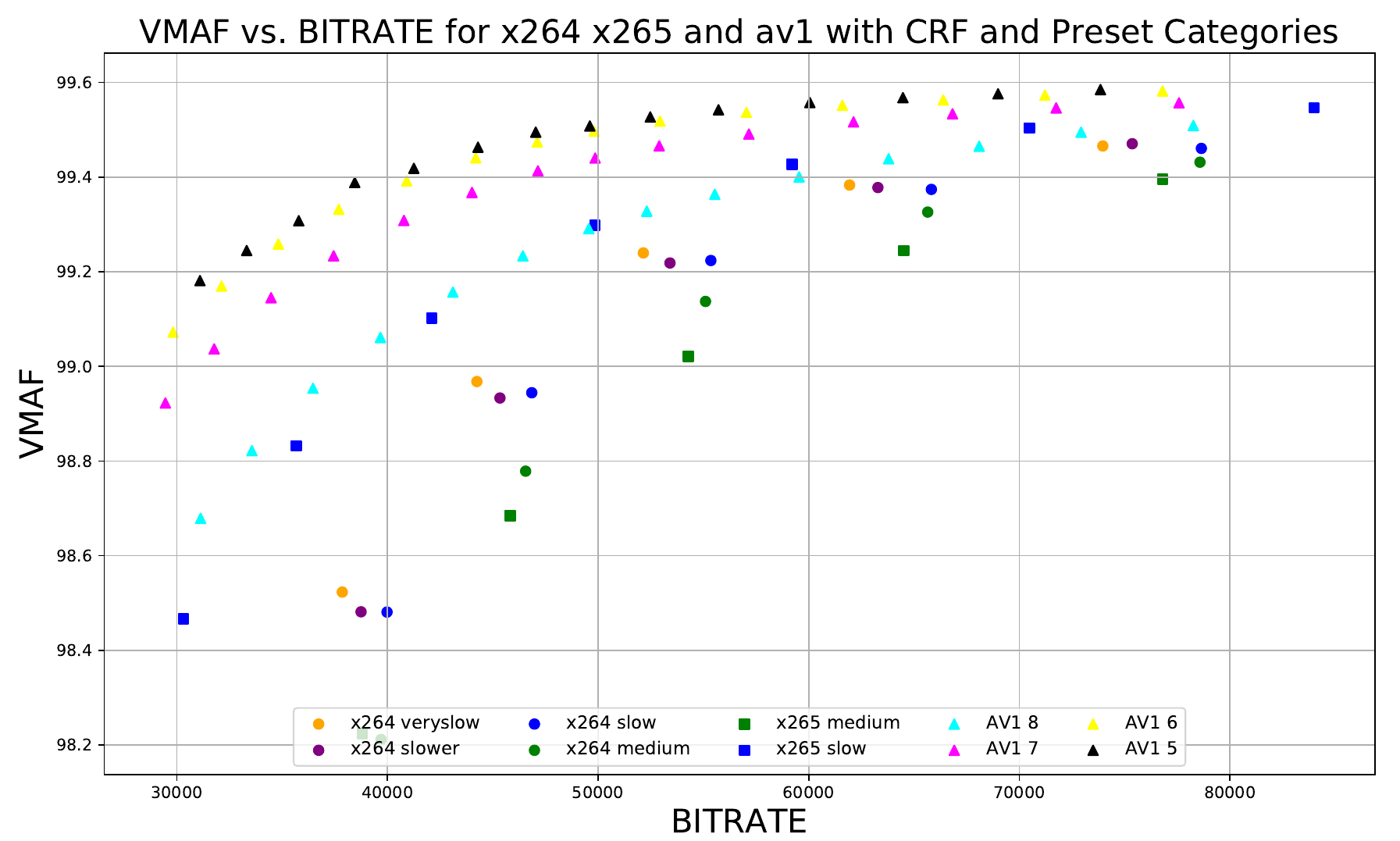}
	\caption{All codecs VMAF vs BITRATE}
	\label{fig:all-BITRATE}
	
\end{figure}

To accurately evaluate video quality, we first converted the selected rosbag clip to YUV444 format, a format commonly used in video editing due to its ability to preserve the original quality of the video well. We used the YUV444 format video as the testing ground truth to accurately measure the quality of the compressed video. To convert from the raw Bayer patter GBRG8 to YUV444 we used the debayering algorithm R and B Interpolation, as presented in \cite{sakamoto1998software},  to get the BGR image matrix, and then used FFmpeg to convert it to the YUV444 format.

The following experiments all utilize CPU encoding on an Intel Xeon(R) Gold 5218R CPU and 96GB of RAM. There are also encoding blocks included in many GPU cards. But those are all geared towards real-time encoding, e.g. for video streaming, and sacrifice quality for speed.
\cite{safin2020hardware} reports that for mobile robotics, if real-time encoding is required, hardware GPU encoders are the best. But they also find that the highest quality encoding is done by software encoding. The aim for this work is to find options to encode videos in very high quality, with no visible compression artifacts. For this we require the VMAF score of the compressed videos to be at least 99.40.

\begin{figure}[t]
	\centering
	\includegraphics[width=\columnwidth]{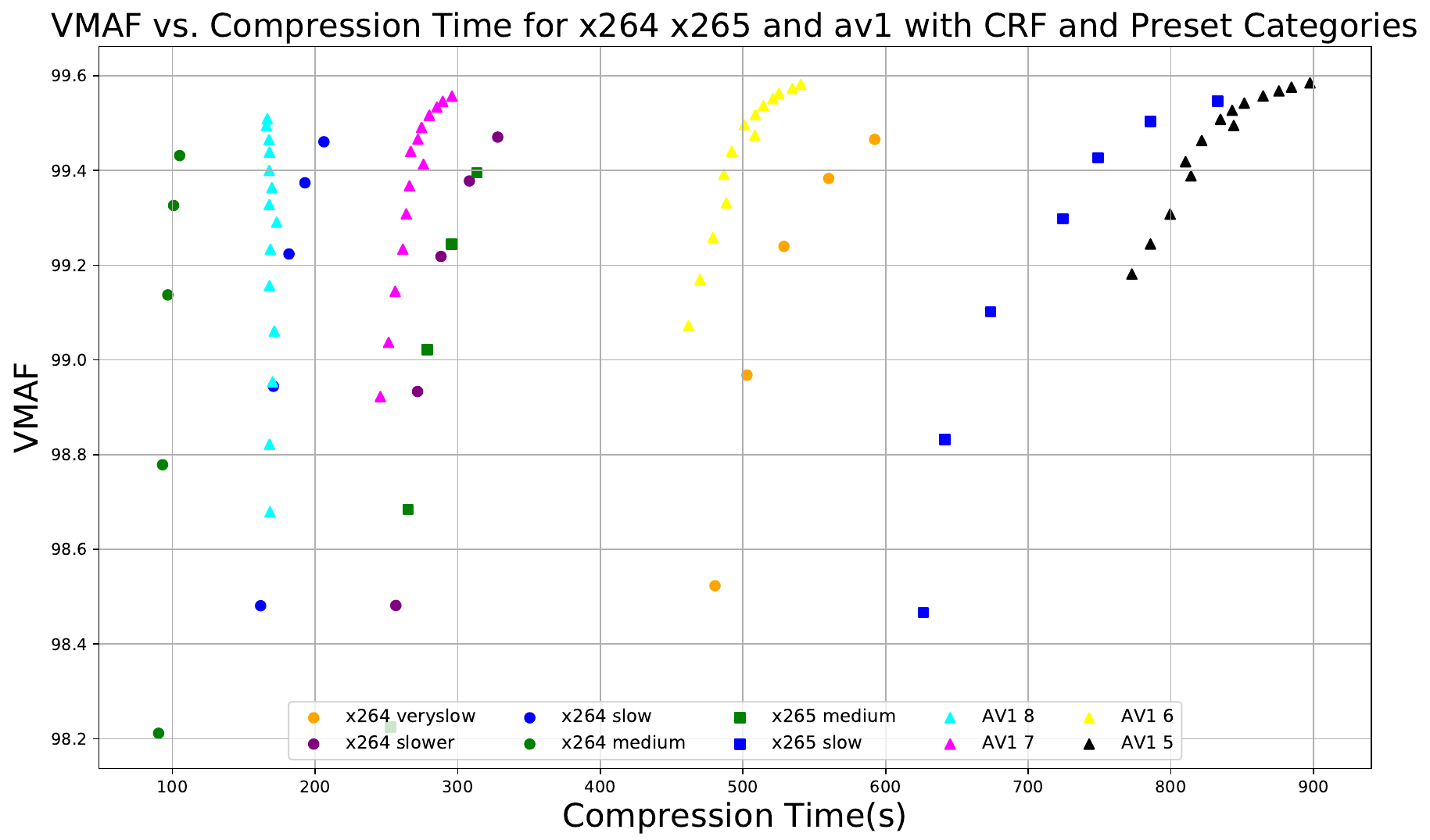}
	\caption{All codecs VMAF vs COMPRESSIONTIME}
	\label{fig:all-time}
	
\end{figure}

In our experiments we selected the CPU codecs x264, x265, and svtav1, and conducted extensive tests using the FFmpeg tool within the range of -crf 16 to 25 and -preset placebo to medium. Tables \ref{tab:Full size}, \ref{tab:libx264-table} and \ref{tab:libx265-table} show the different encoding options and the resulting bitrates, sizes and VMAF of the 60s test video for svtav1, x264 and x265, respectively. The tables are visualized in Fig. \ref{fig:all-BITRATE} and Fig. \ref{fig:all-time} for VMAF vs bitrate and encoding time, respectively. 

 As shown in Fig.~\ref{fig:all-BITRATE}, our test data indicates that svtav1 consistently outperforms the other two codecs. Specifically, the VMAF scores of x264 and x265 only approach 99.50 when the -preset is set to slow and veryslow, while 55\% of svtav1's VMAF scores fall between 99.40 and 99.60, significantly outperforming the other two codecs.

 60s video compression times greater than 1000 seconds would result in compression times of more than 9.7 days for our 14 stream 2h dataset, which exceeds our patience for video post-processing. Therefore, we further narrowed down our analysis to codec settings combinations with VMAF scores between 99.50 and 99.60 and preset values greater than 4, as shown in Table~\ref{tab:Full size}.

Analysis of the Table~\ref{tab:Full size} reveals a compression combination resulting in a good VMAF score: -CRF 19, -preset 6. Given that our primary concern is compression effectiveness and quality, and according to Table~\ref{tab:av1-vision}, there is no visual difference between VMAF scores of 99.50 and 99.55, we ultimately chose the compression combination of -CRF 22 and -preset 6. This combination achieves the best balance in terms of compression effect, quality, and time consumption.

\subsection{Effect of Surce Framerate}
The maximum framerate of our dataset is 60Hz. But occasionally we may also opt to collect lower frame rate data. The question this experiment is answering is: What is the best encoding setting for lower frame rates than 60Hz? The hypothesis is, that higher frame rate videos feature more similar image content, so encoding may be more efficient for those streams. On the other hand, camera data is inherently noisy, such that, at least for very high quality videos, no practical difference may be found.

To answer this question we subsample our one minute test stream by taking only every n-th frame, to get to the frame rates shown in Table \ref{tab:fps}. For a fixed VMAF of about 99.50, we then record the average frame sizes and encoding times per frame.

The results show that indeed higher frame rates encode each frame more efficiently. But, since the settings kind of set a goal bitrate, given our goal VMAF, for lower frame rates, lower settings can be chosen. The table shows good settings for very high quality encoding at lower frame rates with acceptable sizes and encoding times.

\setlength{\tabcolsep}{4pt}

\begin{table}
	\centering
	\caption{Frame rate vs. frame size at about VMAF 99.50}
	\label{tab:fps}
	\begin{tabular}{lrrrrrrr} 
		\hline
		Frame rate & 2Hz & 5Hz& 10Hz & 20Hz & 30Hz & 30Hz &60Hz \\
		Frame size (kb) & 379.1 & 369.4 & 317.3 & 234.4 & 196.7 & 144.2 & 101.3 \\ 
		Frame time (s) & 0.14 & 0.09 & 0.08 & 0.06 & 0.05 & 0.14 & 0.14 \\ 
		Actual VMAF & 99.50 & 99.50 & 99.51 & 99.50 & 99.50& 99.52 & 99.50 \\ 
		Crf & 18 & 17 & 17 & 18 & 18 & 22 & 22 \\ 
		Preset & 8 & 8 & 8 & 8 & 8 & 6 & 6 \\ \hline
	\end{tabular}
\end{table}

\setlength{\tabcolsep}{10pt}

\subsection{Compression Results of "ShanghaiTech Mapping Robot Outdoor Dataset"}
\label{sec:dataset}

The practical test of this paper comes from the Mapping Robot Outdoor Dataset \cite{xu2024shanghaitech}. This is a ground-based unmanned vehicle dataset that includes multiple sensors. In particular, we focus on the recorded data from two types of RGB cameras provided by the dataset: FLIR Grasshopper3 (GS3-U3-51S5C-C) and FLIR LadyBug5+. They provide high-resolution and high-frame-rate recorded data. We tested our dataset scheme on the data collected by these two cameras, as shown Table~\ref{tab:compression result}. For this dataset we ran with half the maximum framerate. The results show that compared with the original data, the dataset scheme using video compression technology can save 2,436.5GB of space.

\begin{table}
	\centering
	\caption{Compression Results in GB of 26min ShanghaiTech\\ Mapping Robot Outdoor Dataset}
	\label{tab:compression result}
	\begin{tabular}{lrr} 
		\hline
		& GS3-U3-51S5C-C & LadyBug5+ \\
		Video streams & 11 & 6 \\
		Frame Rate & 30 Hz & 15 Hz \\ \hline
        Raw Data &2,389&92.0\\
        YUV444 &7,167&276.0\\
         RGB &7,167&276.0\\
         JPG Quality 90&446&17.1\\
         JPG Quality 75&250&9.7\\
         Our Dataset &43&1.7\\
		\hline
	\end{tabular}
\end{table}

\section{Conclusions}
\label{sec:conclusions}

This paper presented a method for encoding robotic dataset video data as mp4 files. Our software allows the replay of mp4 files within the ROS framework, as if the video was stored in a ROS bagfile, including timestamps and replay with simulated time. This enables us to utilize modern video codecs like av1 and is also user friendly, because the data can be viewed with any video player and the data is compatible with ROS 1 and ROS 2 without the need for conversion. The paper furthermore evaluated which codec and setting is the best for very high video quality without any visible compression artifacts, while maintaining reasonable compression size and encoding time and finally selected an av1 setting. This encoding is more than ten times smaller and with higher quality than JPG encoding with relatively high quality 90. We also explored the effect of source frame rate on the encoding and found that higher frame rates encode frames more efficiently due to higher similarity in the frames. With those frame rate experiments we also suggested settings for high-quality encoding with lower source frame rates. In our ShanghaiTech Mapping Robot paper we then applied those results to a real, big robotic dataset.

In the future we expect more datasets from the ShanghaiTech Mapping Robot, even with 60Hz frame, to be released to the public.

\bibliographystyle{unsrt}

\bibliography{ref}

\begin{thebibliography}{10}

\bibitem{cadena2016past}
Cesar Cadena, Luca Carlone, Henry Carrillo, Yasir Latif, Davide Scaramuzza,
  Jos{\'e} Neira, Ian Reid, and John~J Leonard.
\newblock Past, present, and future of simultaneous localization and mapping:
  Toward the robust-perception age.
\newblock {\em IEEE Transactions on robotics}, 32(6):1309--1332, 2016.

\bibitem{liu2024benchmarking}
Xinzhe Liu, Yuanyuan Yang, Bowen Xu, and S{\"o}ren Schwertfeger.
\newblock Benchmarking slam algorithms in the cloud: The slam hive benchmarking
  suite.
\newblock {\em arXiv preprint arXiv:2406.17586}, 2024.

\bibitem{yang2023slam}
Yuanyuan Yang, Bowen Xu, Yinjie Li, and S{\"o}ren Schwertfeger.
\newblock The slam hive benchmarking suite.
\newblock In {\em 2023 IEEE International Conference on Robotics and Automation
  (ICRA)}, pages 11257--11263. IEEE, 2023.

\bibitem{Geiger2012CVPR}
Andreas Geiger, Philip Lenz, and Raquel Urtasun.
\newblock Are we ready for autonomous driving? the kitti vision benchmark
  suite.
\newblock In {\em Conference on Computer Vision and Pattern Recognition
  (CVPR)}, 2012.

\bibitem{chen2020advanced}
Hongyu Chen, Zhijie Yang, Xiting Zhao, Guangyuan Weng, Haochuan Wan, Jianwen
  Luo, Xiaoya Ye, Zehao Zhao, Zhenpeng He, Yongxia Shen, et~al.
\newblock Advanced mapping robot and high-resolution dataset.
\newblock {\em Robotics and Autonomous Systems}, 131:103559, 2020.

\bibitem{xu2024shanghaitech}
Bowen Xu, Xiting Zhao, Delin Feng, Yuanyuan Yang, and S{\"o}ren Schwertfeger.
\newblock Shanghaitech mapping robot is all you need: Robot system for
  collecting universal ground vehicle datasets.
\newblock {\em arXiv preprint arXiv:2406.16713}, 2024.

\bibitem{yang2022cluster}
Yuanyuan Yang, Delin Feng, and S{\"o}ren Schwertfeger.
\newblock Cluster on wheels.
\newblock In {\em 2022 International Conference for Advancement in Technology
  (ICONAT)}, pages 1--8. IEEE, 2022.

\bibitem{chan2023influence}
Pak~Hung Chan, Anthony Huggett, Georgina Souvalioti, Paul Jennings, and
  Valentina Donzella.
\newblock Influence of avc and hevc compression on detection of vehicles
  through faster r-cnn.
\newblock {\em IEEE Transactions on Intelligent Transportation Systems}, 2023.

\bibitem{gummadi2023correlating}
Daniel Gummadi, Pak~Hung Chan, Hetian Wang, and Valentina Donzella.
\newblock Correlating traditional image quality metrics and dnn-based object
  detection: a case study with compressed camera data.
\newblock {\em Authorea Preprints}, 2023.

\bibitem{bayer1976color}
BE~Bayer.
\newblock Color imaging array us patent 3 971 065, 1976.

\bibitem{ramil2018real}
Safin Ramil, Roman Lavrenov, Tatyana Tsoy, Mikhail Svinin, and Evgeni Magid.
\newblock Real-time video server implementation for a mobile robot.
\newblock In {\em 2018 11th International Conference on Developments in
  eSystems Engineering (DeSE)}, pages 180--185. IEEE, 2018.

\bibitem{taguchi2008video}
Hideki Taguchi, Masahiro Iwahashi, and Tetsuya Kimura.
\newblock Video data compression for robot to robot communication.
\newblock In {\em 2008 IEEE International Workshop on Safety, Security and
  Rescue Robotics}, pages 140--145. IEEE, 2008.

\bibitem{nenci2014effective}
Fabrizio Nenci, Luciano Spinello, and Cyrill Stachniss.
\newblock Effective compression of range data streams for remote robot
  operations using h. 264.
\newblock In {\em 2014 IEEE/RSJ International Conference on Intelligent Robots
  and Systems}, pages 3794--3799. IEEE, 2014.

\bibitem{tamhankar2003overview}
Arundhati Tamhankar and KR~Rao.
\newblock An overview of h. 264/mpeg-4 part 10.
\newblock In {\em Proceedings EC-VIP-MC 2003. 4th EURASIP Conference focused on
  Video/Image Processing and Multimedia Communications (IEEE Cat. No.
  03EX667)}, volume~1, pages 1--51. IEEE, 2003.

\bibitem{sullivan2012overview}
Gary~J Sullivan, Jens-Rainer Ohm, Woo-Jin Han, and Thomas Wiegand.
\newblock Overview of the high efficiency video coding (hevc) standard.
\newblock {\em IEEE Transactions on circuits and systems for video technology},
  22(12):1649--1668, 2012.

\bibitem{pourazad2012hevc}
Mahsa~T Pourazad, Colin Doutre, Maryam Azimi, and Panos Nasiopoulos.
\newblock Hevc: The new gold standard for video compression: How does hevc
  compare with h. 264/avc?
\newblock {\em IEEE consumer electronics magazine}, 1(3):36--46, 2012.

\bibitem{patel2015review}
Dhruti Patel, Tarun Lad, and Dharam Shah.
\newblock Review on intra-prediction in high efficiency video coding (hevc)
  standard.
\newblock {\em International Journal of Computer Applications}, 975(8887):12,
  2015.

\bibitem{kim2012block}
Il-Koo Kim, Junghye Min, Tammy Lee, Woo-Jin Han, and JeongHoon Park.
\newblock Block partitioning structure in the hevc standard.
\newblock {\em IEEE transactions on circuits and systems for video technology},
  22(12):1697--1706, 2012.

\bibitem{lin2013motion}
Jian-Liang Lin, Yi-Wen Chen, Yu-Wen Huang, and Shaw-Min Lei.
\newblock Motion vector coding in the hevc standard.
\newblock {\em IEEE Journal of selected topics in Signal Processing},
  7(6):957--968, 2013.

\bibitem{han2021technical}
Jingning Han, Bohan Li, Debargha Mukherjee, Ching-Han Chiang, Adrian Grange,
  Cheng Chen, Hui Su, Sarah Parker, Sai Deng, Urvang Joshi, et~al.
\newblock A technical overview of av1.
\newblock {\em Proceedings of the IEEE}, 109(9):1435--1462, 2021.

\bibitem{joshi2019loop}
Urvang Joshi, Debargha Mukherjee, Yue Chen, Sarah Parker, and Adrian Grange.
\newblock In-loop frame super-resolution in av1.
\newblock In {\em 2019 Picture Coding Symposium (PCS)}, pages 1--5. IEEE, 2019.

\bibitem{tomar2006converting}
Suramya Tomar.
\newblock Converting video formats with ffmpeg.
\newblock {\em Linux journal}, 2006(146):10, 2006.

\bibitem{rassool2017vmaf}
Reza Rassool.
\newblock Vmaf reproducibility: Validating a perceptual practical video quality
  metric.
\newblock In {\em 2017 IEEE international symposium on broadband multimedia
  systems and broadcasting (BMSB)}, pages 1--2. IEEE, 2017.

\bibitem{sakamoto1998software}
Tadashi Sakamoto, Chikako Nakanishi, and Tomohiro Hase.
\newblock Software pixel interpolation for digital still cameras suitable for a
  32-bit mcu.
\newblock {\em IEEE Transactions on Consumer Electronics}, 44(4):1342--1352,
  1998.

\bibitem{safin2020hardware}
Ramil Safin, Emilia Garipova, Roman Lavrenov, Hongbing Li, Mikhail Svinin, and
  Evgeni Magid.
\newblock Hardware and software video encoding comparison.
\newblock In {\em 2020 59th Annual Conference of the Society of Instrument and
  Control Engineers of Japan (SICE)}, pages 924--929. IEEE, 2020.

\end{thebibliography}

\end{document}